\pdfoutput=1

\documentclass[11pt]{article}

\usepackage[final]{acl}
\usepackage{hyperref}
\usepackage{tabularx}
\usepackage{lipsum}  
\usepackage{array}
\usepackage{graphicx}

\usepackage{float}
\usepackage{times}
\usepackage{latexsym}

\usepackage[T1]{fontenc}

\usepackage[utf8]{inputenc}

\usepackage{microtype}

\usepackage{inconsolata}

\usepackage{graphicx}

\usepackage{booktabs}
\usepackage{multirow}
\usepackage{adjustbox}
\usepackage{xspace}
\usepackage{tabularx}
\usepackage{multicol}

\title{Machine Translation Hallucination Detection for Low and High Resource Languages using Large Language Models}

\usepackage{xspace}
\newcommand*{\halomi}{\textit{HalOmi}\xspace}
\newcommand*{\llama}{\texttt{Llama3-70B}\xspace}
\newcommand*{\gpt}{\texttt{GPT4-Turbo}\xspace}
\newcommand*{\gpto}{\texttt{GPT4o}\xspace}
\newcommand*{\crplus}{\texttt{Command R+}\xspace}

\author{
 \textbf{Kenza Benkirane*\textsuperscript{1}},
 \textbf{Laura Gongas*\textsuperscript{1}}, \\
  \textbf{Shahar Pelles\textsuperscript{1}},
 \textbf{Naomi Fuchs\textsuperscript{1}},
 \textbf{Joshua Darmon\textsuperscript{1}}, \\
 \textbf{Pontus Stenetorp\textsuperscript{1}}, \textbf{David Ifeoluwa Adelani\textsuperscript{2,3}}, \textbf{Eduardo Sánchez\textsuperscript{1,4}}
\\
\\
 \textsuperscript{1}University College London, \textsuperscript{2}Mila - Quebec AI Institute
, \textsuperscript{3}McGill University,
 \textsuperscript{4}Meta
 \\
 \small{*: Equal contribution}
\\
}

\begin{document}
\maketitle
\begin{abstract}
Recent advancements in massively multilingual machine translation systems have significantly enhanced translation accuracy; however, even the best performing systems still generate hallucinations, severely impacting user trust. Detecting hallucinations in Machine Translation (MT) remains a critical challenge, particularly since existing methods excel with High-Resource Languages (HRLs) but exhibit substantial limitations when applied to Low-Resource Languages (LRLs). This paper evaluates sentence-level hallucination detection approaches using 
Large Language Models (LLMs) and semantic similarity within massively multilingual embeddings. Our study spans 16 language directions, covering HRLs, LRLs, with diverse scripts. We find that the choice of model is essential for performance. On average, for HRLs, \llama 
outperforms 
the previous state of the art by as much as 0.16 MCC (Matthews Correlation Coefficient). However, for LRLs we observe that \texttt{Claude Sonnet} outperforms other LLMs on average by 0.03 MCC. The key takeaway from our study is that LLMs can achieve performance comparable or even better than previously proposed models, despite not being explicitly trained for any machine translation task. However, their advantage is less significant for LRLs.
\footnote{Data and code are available on \href{https://github.com/kenza-ily/mt_hallucination_detection}{GitHub}.}

\end{abstract}

\section{Introduction}

\begin{figure}[t]
    \centering
    \vspace{-2mm}
    \includegraphics[width=0.95\linewidth]{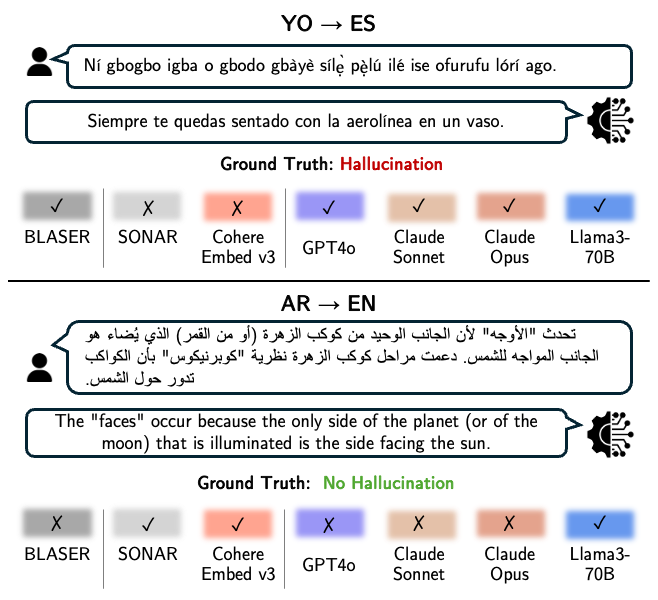}
    \caption{Illustration of how a selection of the evaluated methods perform from Yoruba to Spanish and from Arabic to English.}
    \label{fig:emb_vs_llms}
     \vspace{-2mm}
\end{figure}

Text generation models have drastically improved in recent years especially with the capabilities of LLMs in producing realistic and fluent output. However, hallucination continues to undermine user trust, as it generates and propagates misinformation and sometimes nonsensical outputs~\citep{51844,wu_tacl_a_00563,guerreiro-etal-2023-looking}. This issue is especially critical in high-stakes domains like medicine and law, where hallucinations in medical texts can result in harmful misunderstandings about diagnoses or treatment instructions, and inaccuracies in legal contract translations can lead to severe financial or legal consequences.

One practical way of reducing hallucination in MT is by building more robust models, especially for LRL which tend to exhibit significantly higher hallucination rates. There are several efforts on scaling MT models to LRLs, such as M2M-100~\citep{fan2020englishcentric}, NLLB-200~\cite{Team2022}, MADLAD-400~\cite{kudugunta2023madlad400} etc. Despite initiatives to minimize hallucinations during the MT process, issues still persists. Therefore, detecting hallucinations post-translation remains a critical alternative approach to ensure the reliability and trustworthiness of the translated content.

Previous work on post-translation evaluation has primarily focused on general translation errors, with evaluation scores often under-representing the impact of hallucinations due to their relatively low frequency compared to less severe errors like omissions \cite{guerreiro2023xcomet}. Studies on hallucinations have mainly concentrated on English-centric (\texttt{EN}) to HRL translation direction, while research involving non-English LRLs remains limited \citep{Raunak2022, Xu}.

For instance, sentence similarity measures between source and translated texts using cross-lingual embeddings, such as LASER \cite{Heffernan2022} and LaBSE \cite{Feng2022}, have proven effective at identifying severe hallucinations \cite{Dale2022}, though their limitations with LRLs are often overlooked \cite{Dale2023}. Recent studies have highlighted the capabilities of LLMs in multilingual MT evaluation, demonstrating strong performances across various languages, although discrepancies remain for LRLs \cite{Zhu, Xu}. \citet{Kocmi2023} demonstrated that, when properly prompted, LLMs can assess the quality of machine-generated translations, achieving state-of-the-art results in system-level evaluation for HRLs. Furthermore, \citet{fernandes-etal-2023-devil} pioneered the use of LLMs for MT tasks in LRLs with the introduction of the \textit{AUTOMQM} prompting technique, though this approach primarily targets broader translation errors rather than focusing on hallucination detection.

Recently, \citet{Dale2023} introduced \halomi--- a benchmark dataset for detecting hallucination in MT that includes \texttt{EN$\leftrightarrow$HRLs} (ten directions)  and \texttt{ EN$\leftrightarrow$LRLs} (six directions), as well as two non-English directions \texttt{HRL$\leftrightarrow$LRL}, including different scripts. \texttt{BLASER-QE} \cite{Costa-jussà}, the state-of-the-art (SOTA) hallucination detector, is reported as the top performer on the \halomi benchmark. It calculates a translation quality score by evaluating the similarity between encoded source texts and machine-translated texts within the SONAR embedding space \cite{Duquenne2023}.\\
\indent In this paper, we evaluate the performance of LLMs and embedding based methods as hallucination detectors, aiming to enhance performance in both HRLs and LRLs. To this end, we use the \halomi benchmark dataset with a binary sentence-level hallucination detection approach. For our 
evaluation, we include 14 methods: eight LLMs with different prompt variations, and four embedding spaces by computing the cosine similarity between source and translated texts.\\
\indent We find that LLMs are highly effective for hallucination detection across both high and low resource languages, although the optimal model selection depends on specific contexts. For HRLs, on average across directions, the \texttt{\llama} model significantly surpasses the previous SOTA method, \texttt{BLASER-QE}, by 16 points. Moreover, embedding-based methods have also demonstrated superior performance over the current SOTA in high resource contexts.  However, for LRLs, \texttt{Claude Sonnet} is the best performing model, improving previous methods by a smaller difference. More precisely, LLMs outperformed \texttt{BLASER-QE} in five out of eight LRL translation directions, including the non-English-centric ones. \\
\indent Finally, our research makes the following primary contributions: First, we evaluate a wide range of LLMs for MT hallucination detection and establish that LLMs, despite not being explicitly trained for the task, are competitive and \textit{greatly} outperform even the previous SOTA for HRLs. Second, large multilingual embedding spaces improve upon previously proposed methods and show that they remain competitive for HRLs, but struggle for LRLs. Third, we establish a new SOTA for 13 of the 16 languages that we evaluate on, including high and low resource languages. Surpassing the previous SOTA, which was explicitly trained for the task, on average by 2 MCC points.

\section{Experimental setup}


\subsection{Quality assessment of the dataset}

We evaluated our methods on the \halomi dataset.
A first dataset filtration involved selecting only natural translations, without perturbations, as findings from perturbed data may not be applicable to the detection of natural hallucinations \cite{Dale2023}.

The validation and test split was decided based on the translation direction.
For the validation set, we selected the two translation directions \texttt{DE$\leftrightarrow$EN}, which encompasses \texttt{301} sentences. This choice was made as extensive resources and established benchmarks are available for this language pair \cite{guerreiro2023xcomet}, with the expectation that the models would exhibit generalizability to less frequently used language pairs.
For the test set, the other 16 pairs were used: more precisely, it includes four pairs with English and a HRL (\texttt{EN$\leftrightarrow$AR}
, \texttt{EN$\leftrightarrow$ZH}, \texttt{EN$\leftrightarrow$RU}, \texttt{EN$\leftrightarrow$ES}), three pairs with English and a LRL (\texttt{EN$\leftrightarrow$KS}, \texttt{EN$\leftrightarrow$MN}, and \texttt{EN$\leftrightarrow$YO}), and one non-English HRL-LRL pair (\texttt{ES$\leftrightarrow$YO}). 
The test set includes \texttt{2,558} sentence pairs. This test set excludes six sentence pairs that were removed due to sensitive content flagged and filtered out by LLMs.
A more detailed description of the dataset is available in {\autoref{sec:app_dataset}}.

\subsection{Hallucination detection setting}

We consider two settings: (1) \textbf{Severity ranking} introduced by the authors of \halomi. (2) \textbf{Binary detection}---a new setting we added due to data imbalance and ease of evaluation. 

\paragraph{Severity ranking} the classification of hallucinations was based on four severity levels: \textit{No Hallucination}, \textit{Small Hallucination}, \textit{Partial Hallucination}, and \textit{Full Hallucination}. This fine-grained categorization aimed to capture the nuances in the extent and impact of hallucinations on the translated output.  We use this setting only as \textbf{ablation study} in {\autoref{sec:app_ablation}}., both for consistency with the \halomi benchmark, but also to assess the relevance of our binary detection approach.


\paragraph{Binary detection} 
In this setting, all three instances of hallucinations were labelled as \textit{Hallucination}, regardless of their severity. We also change the way the evaluation was done in \halomi, with an appropriate prompt ({\autoref{sec:app_prompts}}), and threshold calculation for binary classification for embeddings cosine similarity, see {\autoref{sec:exp_embeddings}}. The primary reason for choosing this setting is the significant class imbalance in \halomi, largely due to the scarcity of hallucinations across different severity levels. Some translation directions have particularly imbalanced data, for example \texttt{EN$\rightarrow$RU}, with the following distribution: out of 148 sentence pairs, we have 141 \textit{No Hallucination} (96.6\%), 1 \textit{Small} (0.68\%), 2 \textit{Partial} (1.4\%), and 4 \textit{Full} (2.8\%). High class imbalance can affect the ability of model to perform well~\citep{Prusa2016,Sordo2005,Fernández2013}. 



\subsection{LLMs for hallucination detection}
We assessed the performances of eight LLMs, mixing capabilities models across LLMs families. We evaluate OpenAI's \texttt{GPT4-turbo} and \texttt{GPT4o}; Cohere's \texttt{Command R} and \crplus;   Mistral's \texttt{Mistral-8x22b}; Anthropic's  \texttt{Claude Sonnet} and \texttt{Claude Opus} and Meta's \texttt{\llama}.\footnote{\texttt{GPT3.5}, \texttt{Mistral Large} and \texttt{Llama3-8B} were initially taken into account, but were excluded due to poor task understanding.} More details about the selection are in {\autoref{sec:llm_choice}}.

First, we built our prompt design by differentiated system and user prompts for better results \cite{Kong2023}. The system prompt contained the task description, and optionally, the inclusion of Chain-of-Thought (CoT), while the user prompt contained, for each sentence pair, the source text and MT text, as well as a direct hallucination classification question.

\begin{figure}[t]
    \centering
    \vspace{-2mm}
    \includegraphics[width=0.95\linewidth]{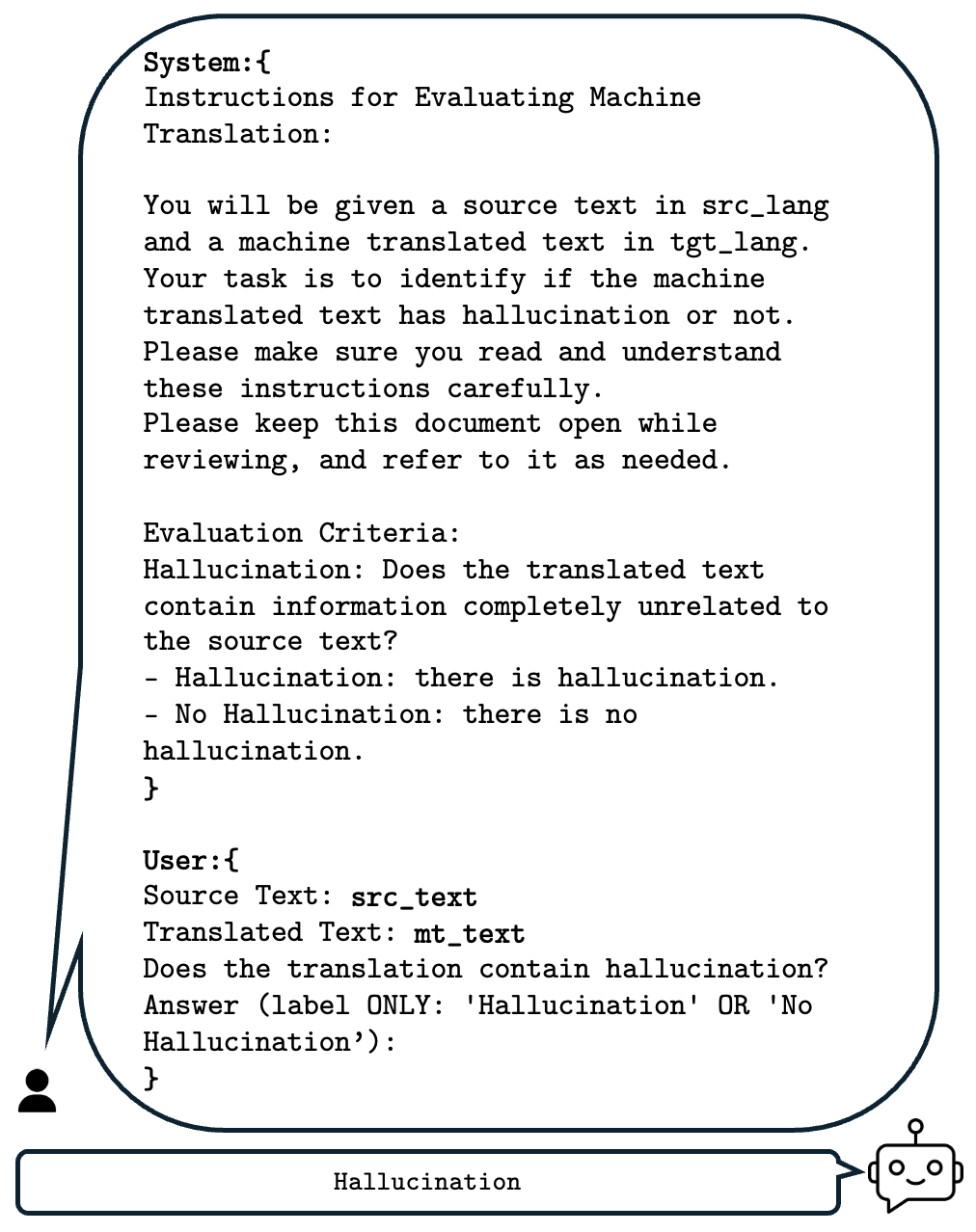}
    \caption{Binary detection prompt sample.}
    \label{fig:botpic}
     \vspace{-2mm}
\end{figure}

We derived the task description prompts from the \textit{Evaluate Hallucination} and \textit{Evaluate Coherence in the Summarization Task} prompts in \textit{G-Eval} \cite{Liu2023}. The CoT prompts were inspired by \textit{Evaluation Steps} from \textit{G-Eval}, and by the human annotation guidelines and severity level definitions from \textit{HalOmi}. All prompts are available {\autoref{sec:app_prompts}}, with \hyperref[fig:botpic]{Figure \ref{fig:botpic}} showing one of the prompts used for binary detection. More details about the chosen hyperparameters with LLMs can be found in {\autoref{sec:app_LLM_exp}}.

We determined the optimal prompts for each model using the \texttt{DE$\leftrightarrow$EN} validation set, evaluating three prompts and two CoT proposals for binary detection. The best prompt for each model was selected based on the average MCC across both translation directions. 
The MCC was chosen as the primary metric for binary detection due to its superiority in providing a single, easily interpretable value between -1 and +1. This value encapsulates the model's performance for the confusion matrix scores, making it more robust to class imbalance.

\subsection{Embeddings}
\label{sec:exp_embeddings}

We assessed the performance of three LLM-related embedding spaces: OpenAI's \texttt{text-embedding-3-large}, Cohere's \texttt{Embed v3}, and Mistral's \texttt{mistral-embed}. Additionally, we included \texttt{SONAR}, the multilingual embedding space used as the base for \texttt{BLASER-QE}. Specifically, we calculated the cosine distance between embeddings of the source text and the machine-translated text. This approach draws on previous studies showing that hallucinated translations tend to have embeddings that are significantly distanced from those of the source text \cite{Dale2022}.

We binarised the cosine similarity scores of embeddings using an optimal threshold value determined from the validation set. This threshold, established by maximizing the F1-score from the precision-recall curve, was then applied to the test set for binary hallucination detection across all language pairs. Each embedding space was independently processed to maintain the integrity of the evaluation.

\section{Results}


\begin{figure*}[t]
\includegraphics[width=1\textwidth]{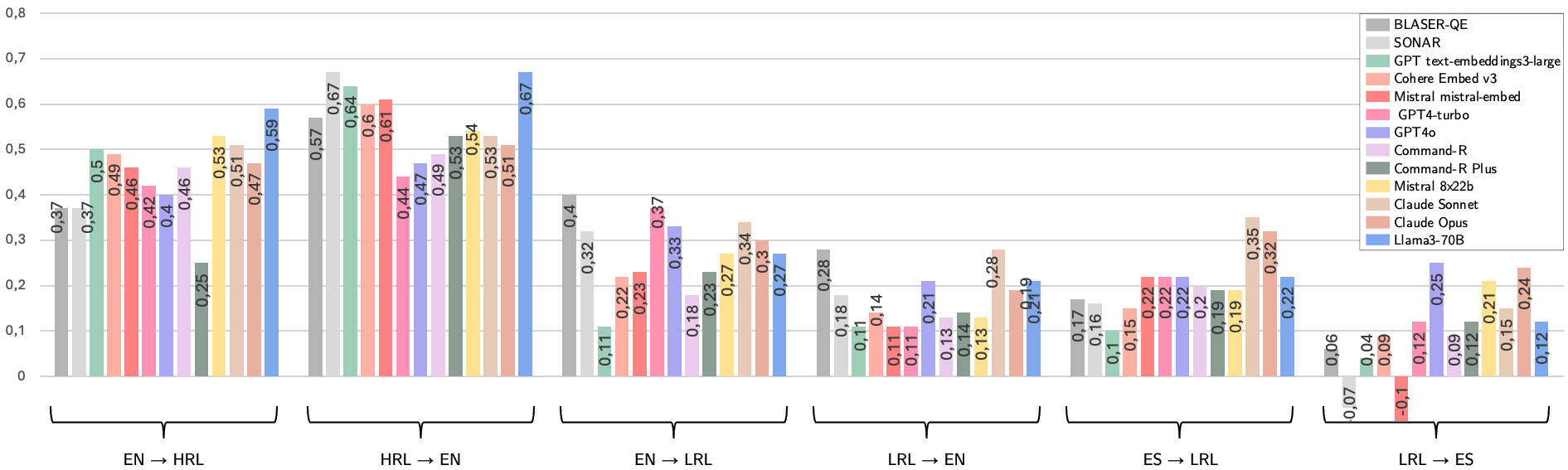}
\caption{MCC average score across high and low resource levels, for different directions. The best performing models differ significantlly between HRLs and LRLs. For HRLs, \llama greatly outperforms other methods, whereas for LRLs, best performers differ from and to LRLs, with \texttt{Claude} and \texttt{GPT} models closely competing. Embeddings demonstrate impressive results, particularly for the \texttt{EN$\rightarrow$HRL} directions.}
\label{fig:binary_test_barplot}
\end{figure*}


\paragraph{LLMs are the new SOTA for hallucination detection} The results in \autoref{fig:binary_test_barplot} and \autoref{fig:results_heatmaps} demonstrate that LLMs have the best overall performance across languages for binary hallucination detection. Specifically, \llama surpasses the previous best performing model, \texttt{BLASER-QE}, by $+5$ points, with an MCC of $0.43$. For HRLs, 10 out of 12 evaluated methods outperform \texttt{BLASER-QE} ($0.46$), with \llama greatly improving over the baseline by 16 points ($0.63$). Notably, the results show that the choice of LLM should rely on the resource level; as for LRLs, \texttt{Claude Sonnet} achieves the highest average MCC. 
However, \texttt{GPT4o} was the more robust LLM across all languages, with the lowest standard deviation.  Finally, for 13 out of the 16 evaluated translation directions, the evaluated methods outperform \texttt{BLASER-QE}, with the exception of \texttt{KS$\rightarrow$EN}, \texttt{YO$\rightarrow$EN} and \texttt{EN$\rightarrow$MNI}. Our findings on LLMs' superior hallucination detection capabilities align with prior research on their effectiveness in MT quality assessment \cite{Kocmi2023}. 


\paragraph{Embedding-based hallucination detectors remain competitive for HRLs} For HRLs, simple embedding-based methods display competitive capabilities, outperforming more sophisticated models in five out of eight translation directions. For instance, although \texttt{BLASER-QE} is a more advanced model based on \texttt{SONAR}, \texttt{SONAR} exhibits comparable or superior performances in most HRLs directions. This suggests that the effectiveness of these methods may be highly sensitive on their training data, and hence to the resource level, as we observe SOTA performances for HRLs and suboptimal results for LRLs. 
Additionally, the embeddings' performance may be highly dependent on the threshold chosen using the \texttt{EN$\leftrightarrow$DE} validation set, generalizing well for HRLs but not for LRLs.


\paragraph{LLMs' contrastive performances across LRLs}
First, while \llama obtains the best performance overall, it was outperformed in most translation directions, especially in LRL. This result reveals a HRLs-centric approach of the model but also concludes that there is not one-LLM fits all resource levels. 
Secondly, for LRLs, models such as \texttt{Sonnet}, \texttt{Opus}, \texttt{GPT4o}, and \texttt{Mistral} —in order of decreasing performances, achieve higher scores, supporting the feasibility of employing LLMs in settings encompassing a wide range of languages. These results should be contrasted with a wide difference of hallucination distribution across resource levels, for example with the \texttt{MN$\rightarrow$EN} direction which only has $28\%$ \textit{No hallucination} sentence pairs.
More precisely, \texttt{Sonnet} and \texttt{BLASER-QE} perform on par for LRL, with the particularity that \texttt{BLASER-QE} has a significantly higher rate of false negatives, while \texttt{Sonnet} maintains a more balanced ratio of false positives to negatives. Moreover, \texttt{BLASER-QE} performs well in translations from English and comparably to \texttt{Sonnet} in translations to English, but falls short in non-English-centric translations, which follows the same trends as previously reported models in \cite{Dale2023}. \autoref{fig:binary_test_barplot} and \autoref{fig:results_heatmaps} provide a more detailed view of these performance metrics.

\paragraph{Embeddings are high performers for non-Latin scripts, while LLMs can generalise to non-English centric translations}
For \texttt{HRLs$\rightarrow$EN} directions with source scripts different than Latin (\texttt{AR,RU,ZH}), embeddings are the best performers, suggesting high capabilities with cross-script transfer learning. These observations align with the findings of \citet{Hada2023}, who report decreased performance for non-Latin scripts in LLM-based evaluators.
In the two non-English centric translation directions (\texttt{ES$\leftrightarrow$YO}), \texttt{Opus} outperforms by far both \texttt{BLASER-QE} ($0.11$) and the best embedding \texttt{Mistral} ($0.12$), with a score  of $0.28$. Unlike the overall LRLs trends, \texttt{Opus} outperforms \texttt{Sonnet} for this direction pair: this can suggest that the advanced analytical capabilities of LLMs can generate improved results even in scenarios with limited relevant training data. 
Remarkably, in the \texttt{YO$\rightarrow$ES} translation direction, six out of our fourteen methods and \texttt{BLASER-QE} exhibit scores close to random guessing (within the [–1, +1] range). This observation underscores the pressing need for enhanced capabilities in detecting hallucinations in non-English-centric translation settings. \autoref{fig:emb_vs_llms} presents two examples that highlight the challenges faced by LLMs when dealing with non-Latin scripts, with the exception of \texttt{Llama3-70B}. Additionally, it illustrates how embeddings may struggle with reasoning capabilities in non-English centric contexts.

\section{Conclusion}
In this work, we demonstrates that LLMs and embedding semantic similarity are highly effective for hallucination detection in machine translation, with LLMs establishing a new state-of-the-art performance across both high and low-resource languages. Our findings suggest that the optimal model selection depends on specific contexts, such as resource level, script, and translation direction. Our study highlights the practical advantage of reference-free models like LLMs, which allow real-time hallucination detection without relying on external knowledge \cite{Su2024,Zhang2023}. However, there remains a critical need for further research to improve hallucination detection, particularly in low-resource and non-English-centric translation settings.

\clearpage

\section*{Limitations}

Despite the promising results obtained by LLMs and embedding-based methods in our evaluation, there are certain limitations that should be noted.

First, the dataset shows distribution imbalance across translation directions, with different trends for high and low resource languages, even after binarisation (see {\autoref{sec:app_dataset}}):
The HRLs show a pronounced data imbalance towards \textit{No Hallucination} labels, with distribution between $79\%$ and $94\%$. Moreover, for LRLs, there's a  broader interval, from $28\%$ to $85\%$. 
This imbalance often results in models that classify translations as \textit{No hallucination} being more frequently correct for HRLs than for LRLs, thereby introducing a bias into the binary evaluation.
Moreover, the translation direction display a qualitative bias, as shown {\autoref{sec:selection}}: HRLs and LRLs don't have the same selection distribution which display a potential bias towards hallucination.
Future dataset improvements should prioritize larger, more diverse samples, non-Latin scripts, and non-English centric translations. Using consistent source text across languages and balancing hallucination severity levels would enable more sophisticated methods, improve generalizability, and allow for a fair evaluation of models' hallucination detection capabilities.

It is important to note there is a possibility of test set contamination, which is a common challenge in LLM research when the full training data is not publicly available. This issue primarily affects HRLs, where LLMs are predominantly trained; therefore, the impact on LRLs performance is expected to be minimal.

The validation set used to identify the optimal threshold for non-LLM methods and the best prompt for LLMs only included \texttt{EN$\leftrightarrow$DE} translations. To improve parameter optimization and generalization across various translation directions, especially for LRLs, cross-validation is recommended for future research, as suggested by \citet{Dale2023} and initially planned for our study. However, financial constraints associated with benchmarking non-open source models prevented the implementation of this approach. Future work should focus on developing novel approaches that excel on well-studied HRLs while generalizing effectively to LRLs, assessing robustness, or exploring alternative methods to address this challenge within the limitations of dataset size.

Finally, for benchmarking purposes, only the previous state-of-the-art was included for comparison against the newly evaluated methods. Therefore, for a more comprehensive analysis, it is recommended to include additional methods previously evaluated by \halomi. Moreover, the benchmark can be further strengthened by identifying fine-grained hallucination spans to enhance interpretability.




\section*{Acknowledgements}
We gratefully acknowledge OpenAI for providing API credits through their Researcher Access API program to Masakhane for evaluating GPT-4 LLMs. We also extend our thanks to Cohere for offering API credits for the use of their LLMs. David Adelani is supported by Canada CIFAR AI Chair program.


\clearpage
\bibliography{reference}

\clearpage
\appendix

\section{Dataset description}
\label{sec:app_dataset}

\subsection{Language acronyms mapping}

The languages acronyms follow this mapping throughout the paper: Arabic (\texttt{AR}), Chinese (\texttt{ZH}), English (\texttt{EN}), German (\texttt{DE}), Kashmiri (\texttt{KA}), Manipuri (\texttt{MN}), Russian (\texttt{RU}), Spanish (\texttt{ES}), and Yoruba (\texttt{YO}).

\subsection{Hallucination distribution}
\subsubsection{Distribution of Hallucination in the severity ranking framework}

{\tiny 
\noindent\begin{tabularx}{\columnwidth}{@{}lXXXX@{}}
\toprule
Direction  \newline \textbf{Total} & \textit{1 No} & \textit{2 Small} & \textit{3 Partial} & \textit{4 Full} \\
\midrule
\texttt{DE$\rightarrow$EN} \newline \textbf{155} & 140 \newline \textit{90.32\% }& 2 \newline \textit{1.29\%} & 2 \newline \textit{1.29\%} & 11 \newline \textit{7.10\%}  \\
\texttt{EN$\rightarrow$DE} \newline \textbf{146} & 132 \newline \textit{90.41\%} & 3 \newline \textit{2.05\%} & 2 \newline \textit{1.37\%} & 9 \newline \textit{6.16\%}  \\
\textbf{Total} \newline \textbf{301} & 272 \newline 68.25\% & 5 \newline 1.25\% & 4 \newline 1.00\% & 20 \newline 5.01\% \\
\bottomrule
\end{tabularx}
}

\vspace{1em} 

\begin{table}[ht]
\centering
{\tiny
\noindent\begin{tabularx}{\columnwidth}{@{}lXXXX@{}}
\toprule
Direction \newline \textbf{Total}& \textit{1 No} & \textit{2 Small} & \textit{3 Partial} & \textit{4 Full} \\
\midrule
\texttt{EN$\rightarrow$AR} \newline \textbf{144} & 136 \newline \textit{94.44\%} & 2 \newline \textit{1.39\%} & 2 \newline \textit{1.39\%} & 4 \newline \textit{2.78\%}  \\
\texttt{AR$\rightarrow$EN} \newline \textbf{156} & 132 \newline \textit{84.62\% }& 5 \newline \textit{3.21\%} & 2 \newline \textit{1.28\%} & 17 \newline \textit{10.90\% }  \\
\texttt{EN$\rightarrow$RU} \newline \textbf{146}& 141 \newline \textit{96.58\%} & 1 \newline \textit{0.68\%} & 2 \newline \textit{1.37\%} & 2 \newline \textit{1.37\%}  \\
\texttt{RU$\rightarrow$EN }\newline \textbf{158} & 146 \newline \textit{92.41\%} & 3 \newline \textit{1.90\%} & 2 \newline \textit{1.27\%} & 7 \newline \textit{4.43\%}  \\
\texttt{EN$\rightarrow$ES} \newline \textbf{153} & 131 \newline \textit{85.62\% }& 8 \newline \textit{5.23\%} & 3 \newline \textit{1.96\%} & 11 \newline \textit{7.19\%}  \\
\texttt{ES$\rightarrow$EN} \newline \textbf{160} & 127 \newline \textit{79.38\%} & 17 \newline \textit{10.63\% }& 4 \newline \textit{2.50\%} & 12 \newline \textit{7.50\%}  \\
\texttt{EN$\rightarrow$ZH} \newline \textbf{160} & 131\newline \textit{81.88\%} & 5 \newline \textit{3.13\%} & 4 \newline \textit{2.50\%} & 20 \newline \textit{12.50\%} \\
\texttt{ZH$\rightarrow$EN} \newline\textbf{159} & 127 \newline \textit{79.87\%} & 9 \newline \textit{5.66\%} & 7 \newline \textit{4.40\%} & 16 \newline \textit{10.06\%} \\
\hline
\texttt{EN$\rightarrow$KA} \newline \textbf{184} & 111 \newline \textit{60.33\%} & 8 \newline \textit{4.35\%} & 30 \newline \textit{16.30\%} & 35 \newline \textit{19.02\%}  \\
\texttt{KA$\rightarrow$EN} \newline \textbf{151} & 89 \newline \textit{58.94\%} & 15 \newline \textit{9.93\%} & 32 \newline \textit{21.19\% }& 15 \newline \textit{9.93\%}  \\
\texttt{EN$\rightarrow$YO} \newline \textbf{195} & 166 \newline \textit{85.13\%} & 4 \newline \textit{2.05\%} & 11 \newline \textit{5.64\% }& 14 \newline \textit{7.18\%}  \\
\texttt{YO$\rightarrow$EN} \newline \textbf{146} & 124 \newline \textit{84.93\%} & 4 \newline \textit{2.74\%} & 10 \newline \textit{6.85\%} & 8 \newline \textit{5.48\%}  \\
\texttt{EN$\rightarrow$MN} \newline \textbf{197} & 78 \newline \textit{39.59\%} & 52 \newline \textit{26.40\%} & 54 \newline \textit{27.41\%} & 13 \newline \textit{6.60\% } \\
\texttt{MN$\rightarrow$EN} \newline \textbf{152} & 43 \newline \textit{28.29\% }& 45 \newline \textit{29.61\%} & 58 \newline \textit{38.16\%} & 6 \newline \textit{3.95\%}  \\
\texttt{ES$\rightarrow$YO}\newline \textbf{151} & 97 \newline \textit{64.24\%} & 16 \newline \textit{10.60\% }& 29 \newline \textit{19.21\%} & 9 \newline \textit{5.96\%}  \\
\texttt{YO$\rightarrow$ES} \newline \textbf{152} & 80 \newline \textit{52.63\% }& 26 \newline \textit{17.11\%} & 37 \newline \textit{24.34\%} & 9 \newline \textit{5.92\%}  \\
\textbf{Total} \newline \textbf{2564} & 1859 \newline \textit{72.47\%} & 220 \newline \textit{8.58\%} & 287 \newline \textit{11.19\%} & 198 \newline \textit{7.72\%}  \\
\bottomrule
\bottomrule
\end{tabularx}
}
{\small
\caption{Although fine-grained severity ranking is advantageous for most applications, the rarity of occurrences within each hallucination category may lead to results that lack significance and generalizability due to constrained sample sizes. Notably, within the \textit{HalOmi} dataset, 11 of the 18 language directions include fewer than five samples in at least one hallucination category. To address this limitation, we propose a shift toward binary hallucination detection, where all instances of hallucinations are classified as such, irrespective of their severity. This approach enhances the robustness of the analysis and the significance of results while still evaluating the model's ability to separate even \textit{Small hallucination }(one word in a sentence) from \textit{No hallucinations.}}
}
\label{tab:data_ranking}
\end{table}

\subsubsection{Distribution of Hallucination in the binary detection framework}

\begin{table}[ht]
\centering
{\tiny 
\noindent\begin{tabularx}{\columnwidth}{@{}lXXXX@{}}
\toprule
Direction  & \textbf{Total} & \textit{0 No Hallucination} & \textit{1 Hallucination} \\
\midrule
\texttt{DEU$\rightarrow$EN}  & \textbf{155} & 140 \newline \textit{90.32\% }& 15 \newline \textit{9.68\%} \\
\texttt{EN$\rightarrow$DE} & \textbf{146} & 132 \newline \textit{90.41\%} & 14 \newline \textit{10.00\%}  \\
\textbf{Total} & \textbf{301} & 272 \newline 68.25\% & 29 \newline 31.75\% \\
\bottomrule
\end{tabularx}
}
{\small
\caption{\textbf{Validation set distribution for binary detection, across translation directions, for HRLs and LRLS}}
}
\label{tab:val_set_distribution}
\end{table}

\vspace{1em} 

\begin{table}[ht]
\centering
{\tiny
\noindent\begin{tabularx}{\columnwidth}{@{}lXXX@{}}
\toprule
Direction \newline \textbf{Total} & \textit{0 No Hallucination} & \textit{1 Hallucination} \\
\midrule
\texttt{EN$\rightarrow$AR} \newline \textbf{144} & 136 \newline \textit{94.44\%} & 8 \newline \textit{5.56\%} \\
\texttt{AR$\rightarrow$EN} \newline \textbf{156} & 132 \newline \textit{84.62\%} & 24 \newline \textit{15.38\%} \\
\texttt{EN$\rightarrow$RU} \newline \textbf{146} & 141 \newline \textit{96.58\%} & 5 \newline \textit{3.42\%} \\
\texttt{RU$\rightarrow$EN} \newline \textbf{158} & 146 \newline \textit{92.41\%} & 12 \newline \textit{7.59\%} \\
\texttt{EN$\rightarrow$ES} \newline \textbf{153} & 131 \newline \textit{85.62\%} & 22 \newline \textit{14.38\%} \\
\texttt{ES$\rightarrow$EN} \newline \textbf{160} & 127 \newline \textit{79.38\%} & 33 \newline \textit{20.63\%} \\
\texttt{EN$\rightarrow$ZH} \newline \textbf{160} & 131 \newline \textit{81.88\%} & 29 \newline \textit{18.13\%} \\
\texttt{ZH$\rightarrow$EN} \newline \textbf{159} & 127 \newline \textit{79.87\%} & 32 \newline \textit{20.13\%} \\
\hline
\texttt{EN$\rightarrow$KA} \newline \textbf{184} & 111 \newline \textit{60.33\%} & 73 \newline \textit{39.67\%} \\
\texttt{KA$\rightarrow$EN} \newline \textbf{151} & 89 \newline \textit{58.94\%} & 62 \newline \textit{41.06\%} \\
\texttt{EN$\rightarrow$YO} \newline \textbf{195} & 166 \newline \textit{85.13\%} & 29 \newline \textit{14.87\%} \\
\texttt{YO$\rightarrow$EN} \newline \textbf{146} & 124 \newline \textit{84.93\%} & 22 \newline \textit{15.07\%} \\
\texttt{EN$\rightarrow$MN} \newline \textbf{197} & 78 \newline \textit{39.59\%} & 119 \newline \textit{60.41\%} \\
\texttt{MN$\rightarrow$EN} \newline \textbf{152} & 43 \newline \textit{28.29\%} & 109 \newline \textit{71.71\%} \\
\texttt{ES$\rightarrow$YO} \newline \textbf{151} & 97 \newline \textit{64.24\%} & 54 \newline \textit{35.76\%} \\
\texttt{YO$\rightarrow$ES} \newline \textbf{152} & 80 \newline \textit{52.63\%} & 72 \newline \textit{47.37\%} \\
\textbf{Total} \newline \textbf{2564} & 1859 \newline \textit{72.47\%} & 705 \newline \textit{27.53\%} \\
\bottomrule
\end{tabularx}
}
\caption{\textbf{Testing set distribution for binary detection, across translation directions, for HRLs and LRLS}}
\label{tab:merged_hallucination}
\end{table}

\subsection{Selection distribution}
\label{sec:selection}

\begin{figure}[ht]
    \centering
    \includegraphics[width=1\linewidth]{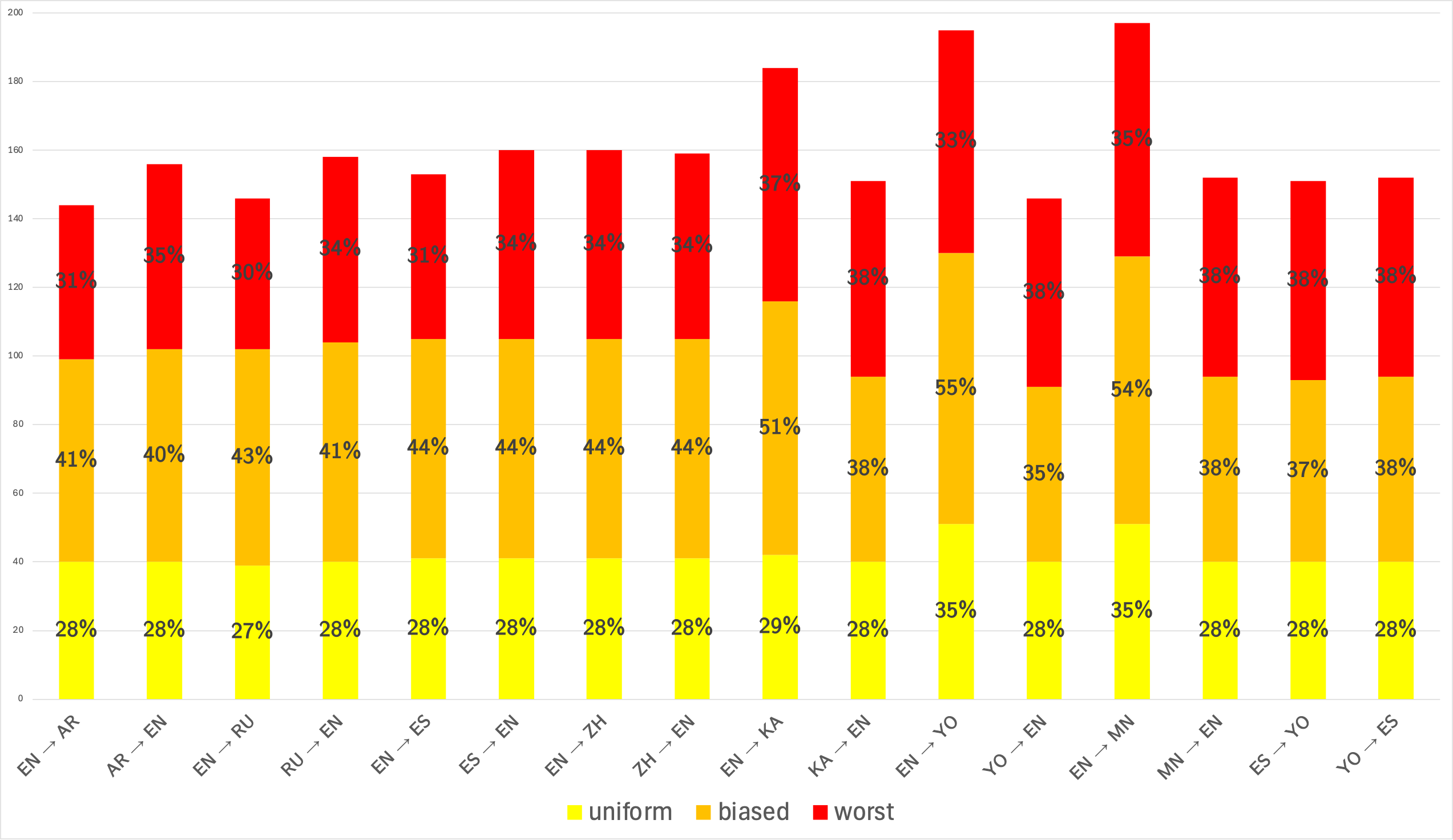}
    \caption{\textbf{Selection type distribution} This graph shows that the three \texttt{EN$\rightarrow$LRLs} not only have more sentences, but also have way more \textit{biased} sentences than other diretions, which suggests a higher propensity to hallucinate.}
    \label{fig:selection}
\end{figure}

The selection information from the \halomi dataset indicates the sampling strategy used to select sentence pairs for each translation direction and data source, which includes \textit{uniform} sampling to maintain data diversity, \textit{biased} sampling favoring potentially problematic translations based on detector quantiles, and \textit{worst} sampling, according to the detectors to increase the likelihood of capturing hallucinations. A closer look at the selection distribution is available {\autoref{fig:selection}}

\section{Ablation study}
\label{sec:app_ablation}
The ablation study focus on hallucination \textbf{severity ranking}. We present results for comparability with \citet{Dale2023}, which assesses the methods' abilities to accurately rank hallucinations by severity (e.g.,full hallucinations ranked higher than partial ones, and any hallucinations ranked above non-hallucinations). The employed metric is an adaptation of the ROC AUC for multiclass tasks, which calculates the percentage of incorrectly ranked sentence pairs with different labels and subtracts this value from the perfect score of 1. We compute these metrics separately for each translation direction to assess the detector's performance across different language pairs.

\subsection{LLMs for severity ranking}

\subsubsection{Prompt design}
We designed tailored prompts for this approach, just as for our main binary approach, this time to generate multiclass predictions. For severity ranking, each prompt has a different assigned CoT.

\paragraph{Severity ranking, Prompt1: \textit{G-Eval}  inspired}

See {\autoref{fig:ranking_prompt1}}

\begin{figure}[ht]
    \centering
    \begin{tabular}{|p{0.9\linewidth}|}
    \hline
    \textit{System:}\\
    \textbf{Human Evaluation of Machine Translation Systems:}

    \textbf{Hallucination Evaluation Criteria:} Does the translated text contain information unrelated to the source text?
    \begin{itemize}
        \item \textbf{No hallucination:} The translated text does not contain any hallucinated words.
        \item \textbf{Small hallucination:} The translated text contains 1-2 hallucinated words.
        \item \textbf{Partial hallucination:} The translated text includes at least 3 hallucinated words, but not all words are hallucinated.
        \item \textbf{Full hallucination:} Nearly all words in the translated text are hallucinated, with the exception of perhaps 1-2 words.
    \end{itemize}
    \textbf{Note:} The labels are mutually exclusive; for example, a translation with a partial hallucination does not qualify as a full hallucination.:

    \textit{User:}\\
    Source Text: \texttt{src\_text} \\
Translated Text: \texttt{mt\_text} \\
Respond with the hallucination level label ONLY. Answer (Labels=[‘No hallucination’,‘Small hallucination’,‘Partial hallucination’,'Full hallucination’])
    \end{tabular}
    \caption{\textbf{Severity Ranking Prompt 1} - from \textit{G-Eval}}
    \label{fig:ranking_prompt1}
\end{figure}

\paragraph{Severity ranking, Prompt2: \textit{G-Eval }inspired, with \textit{HalOmi}'s hallucination deifnition}

See {\autoref{fig:ranking_prompt2}}
\begin{figure}[ht]
    \centering
    \begin{tabular}{|p{0.9\linewidth}|}
    \hline
    \textit{System:}\\
    You will be given a 'Source Text' and its translation, the 'Translated Text'. Your task is to identify the hallucination level of the translated text.

    \textbf{Definition of Hallucination:} \\
    A word in the translated text is considered a hallucination if it introduces information that is completely unrelated to the source text.

    \textbf{Hallucination labels:} 
    \begin{itemize}
        \item \textbf{No hallucination:} The translated text does not contain any hallucinated words.
        \item \textbf{Small hallucination:} The translated text contains 1-2 hallucinated words.
        \item \textbf{Partial hallucination:} The translated text includes at least 3 hallucinated words, but not all words are hallucinated.
        \item \textbf{Full hallucination:} Nearly all words in the translated text are hallucinated, with the exception of perhaps 1-2 words.
    \end{itemize}
    \textbf{Note:} The labels are mutually exclusive; for example, a translation with a partial hallucination does not qualify as a full hallucination.:

    \textit{User:}\\
    Source Text: \texttt{src\_text} \\
    Translated Text: \texttt{mt\_text} \\
    Provide exactly one of the following hallucination level labels as your response. Do not include any additional text or explanation:
    \begin{itemize}
        \item No hallucination
        \item Small hallucination
        \item Partial hallucination
        \item Full hallucination
    \end{itemize}
    \end{tabular}
    \caption{\textbf{Severity Ranking Prompt 2 }-  from \textit{G-Eval} with Hallucination definition}
    \label{fig:ranking_prompt2}
\end{figure}

\paragraph{Severity ranking, Prompt3: \textit{G-Eval} inspired, with \textit{HalOmi}'s hallucination deifnition, and language precision}

See {\autoref{fig:ranking_prompt3}}

\begin{figure}[!ht]
    \centering
    \begin{tabular}{|p{0.9\linewidth}|}
    \hline
    \textit{System:}\\
    You will be given a 'Source Text' in \texttt{src\_lang} and its translation in \texttt{tgt\_lang}, the 'Translated Text'. Your task is to identify the hallucination level of the translated text. Please make sure you read and understand these instructions carefully. Please keep this document open while reviewing, and refer to it as needed.

    \textbf{Definition of Hallucination:} \\
    A word in the translated text is considered a hallucination if it introduces information that is completely unrelated to the source text.

    \textbf{Hallucination labels:} 
    \begin{itemize}
        \item \textbf{No hallucination:} The translated text does not contain any hallucinated words.
        \item \textbf{Small hallucination:} The translated text contains 1-2 hallucinated words.
        \item \textbf{Partial hallucination:} The translated text includes at least 3 hallucinated words, but not all words are hallucinated.
        \item \textbf{Full hallucination:} Nearly all words in the translated text are hallucinated, with the exception of perhaps 1-2 words.
    \end{itemize}
    \textbf{Note:} The labels are mutually exclusive; for example, a translation with a partial hallucination does not qualify as a full hallucination.:

    \textit{User:}\\
    Source Text: \texttt{src\_text} \\
    Translated Text: \texttt{mt\_text} \\
    Provide exactly one of the following hallucination level labels as your response. Do not include any additional text or explanation:
    \begin{itemize}
        \item No hallucination
        \item Small hallucination
        \item Partial hallucination
        \item Full hallucination
    \end{itemize}
    \end{tabular}
    \caption{\textbf{Severity Ranking Prompt 3 }-  from \textit{G-Eval} with Hallucination definition and language precision}
    \label{fig:ranking_prompt3}
\end{figure}

\paragraph{Chain of Thoughts for severity ranking}
See {\autoref{fig:ranking_cot1}} and {\autoref{fig:ranking_cot2}}
\begin{figure}[ht]
    \centering
    \begin{tabular}{|p{0.9\linewidth}|}
    \hline
    \textbf{Evaluation Steps:}
    \begin{enumerate}
        \item Read the source text and the translated text carefully.
        \item To decide whether the translated text contains hallucinations check if the source word “corresponds” to erroneous target tokens. For each work answer:
        \begin{itemize}
            \item Does this source word fall into the common meaning category as this target word?
            \item Does this source word have a semantic connection with this target word?
            \item Can you try to come up with a reasonable theory on how this source word is associated with this target word?
        \item If “no” to all the questions above, then hallucination. Keep a count of the number of hallucinated words for each sentence pair.
        \end{itemize}
        \item After reading all the source and translated text, assign a label to the pair based on the number of hallucinated words.
    \end{enumerate}
    \end{tabular}
    \caption{\textbf{Severity Ranking CoT 1} - from \textit{HalOmi}'s human guidelines}
    \label{fig:ranking_cot1}
\end{figure}

\begin{figure}[ht]
    \centering
    \begin{tabular}{|p{0.9\linewidth}|}
    \hline
    \textbf{Evaluation Steps:}
    \begin{enumerate}
        \item Read the source text and the translated text carefully.
        \item Initialize a counter `n = 0` for the number of hallucinated words.
        \item For each word in the translated text, perform the following checks to determine if it is a hallucinated word:
        \begin{itemize}
            \item Does this source word fall into the common meaning category as this target word?
            \item Does this source word have a semantic connection with this target word?
            \item Can you try to come up with a reasonable theory on how this source word is associated with this target word?
        \item  If "no" to all the questions above, then it is considered a hallucination. Increment `n` by 1.
        \end{itemize}
        \item After analyzing each word in the translated text:
\begin{itemize} 
\item If `n == 0`, assign the label 'No hallucination'.
\item If `n` is 1 or 2, assign the label 'Small hallucination'.
\item If `n` is 3 or more but not all words are hallucinated, assign the label 'Partial hallucination'.
\item If nearly all words are hallucinated, assign the label 'Full hallucination'.'
\end{itemize}
    \end{enumerate}
    \end{tabular}
    \caption{\textbf{Severity Ranking CoT 2} - counting the number of hallucinated words}
    \label{fig:ranking_cot2}
\end{figure}

\subsubsection{Prompt evaluation}
We evaluated three prompts and two CoT variations on the validation set to select the best prompt ({\autoref{tab:val_results_ranking}}). The prompt that achieved the highest average ROC AUC for both directions (\texttt{DE$\leftrightarrow$EN}) was chosen for each method. Subsequently, in the testing phase, each model was assessed with its optimal prompt.

\subsection{Embeddings for severity ranking}
We computed the cosine similarity between the source text and machine-translated text embeddings for each embedding space and took the negative of these results. This approach ensures that hallucinations (indicative of embeddings that are farther apart) correspond to higher numbers, consistent with the ranking scale used in hallucination evaluation. Since this method does not require parameter tuning, the validation set was not utilized for thresholding in contrast to the binary approach.

\subsection{Results}

In the same way as in the binary detection setting, the validation results {\autoref{tab:val_results_ranking}} allowed to select the otpimal prompt for each LLM, and then evaluate this best prompt across the test set, using here the ROC AUC score.
Testing results aredisplayed {\autoref{tab:ranking_results}}, and presents ROC AUC scores for all methods per translation direction. For HRLs, embeddings' high performance remains consistent with the binary hallucination approach.  However, \texttt{BLASER-QE} remains the state-of-the-art in overall performance for severity ranking. The generalizability of these results requires further evaluation due to significant class imbalances in the dataset. Notably, in 11 of the 18 language directions, fewer than five samples are present in at least one hallucination severity category, see {\autoref{sec:app_dataset}}.


\begin{table*}[t]
 \footnotesize
 \begin{center}
 \resizebox{0.7\textwidth}{!}{%
  \begin{tabular}{lrr|rr|rr||rr}
    \toprule
     & \multicolumn{2}{c}{\textit{Prompt1}} & \multicolumn{2}{c}{\textit{Prompt2}} & \multicolumn{2}{c}{\textit{Prompt3}} & \multicolumn{2}{c}{\textit{AVG}} \\
    \textbf{Model} & \textbf{no CoT} & \textbf{CoT1} & \textbf{no CoT} & \textbf{CoT2} & \textbf{no CoT} & \textbf{CoT2} & \textbf{Mean} & \textbf{Std.} \\
    \bottomrule
    \gpt & 0.78 & 0.70 & \textbf{0.83} & 0.81 & 0.82 & 0.81 & 0.79 & 0.05 \\
    \gpto & 0.81 & 0.82 & 0.83 & \textbf{0.83} & 0.83 & 0.83 & \textbf{0.83} & 0.01 \\
    \texttt{Command R} & \textbf{0.82} & 0.79 & 0.77 & 0.80 & \textbf{0.83} & 0.79 & 0.80 & 0.02 \\
    \crplus & 0.75 & 0.76 & 0.79 & \textbf{0.80} & 0.77 & 0.75 & 0.77 & 0.02 \\
    \texttt{Mistral 8x22b} & 0.57 & 0.58 & \textbf{0.77} & 0.67 & 0.69 & 0.67 & 0.66 & 0.07 \\
    \texttt{Sonnet} & \textbf{0.82} & \textbf{0.83} & 0.82 & 0.79 & 0.81 & \textbf{0.83} & 0.82 & 0.01 \\
    \texttt{Opus} & 0.79 & 0.80 & 0.82 & 0.83 & \textbf{0.85} & 0.77 & 0.81 & 0.03 \\
    \llama & 0.80 & 0.81 & \textbf{0.81} & 0.78 & 0.81 & 0.78 & 0.80 & 0.01 \\
    \bottomrule
    \bottomrule
  \end{tabular}
  }
  \vspace{-3mm}
  \caption{Validation results for hallucination detection across prompt variations for severity ranking.}
  \vspace{-4mm}
  \label{tab:val_results_ranking}
  \end{center}
\end{table*}


\begin{table*}[!htbp]
 \footnotesize
 \begin{center}
 \resizebox{\textwidth}{!}{%
  \begin{tabular}{lrrrr|rrrr|rrr|rrr|r|r||rrr}
    \toprule
     & \multicolumn{4}{c}{\texttt{EN$\rightarrow$HRL}} & \multicolumn{4}{c}{\texttt{HRL$\rightarrow$EN}} & \multicolumn{3}{c}{\texttt{EN$\rightarrow$LRL}}
     & \multicolumn{3}{c}{\texttt{LRL$\rightarrow$EN}} & \multicolumn{1}{c}{\texttt{ES$\rightarrow$YO}} & \multicolumn{1}{c}{\texttt{YO$\rightarrow$ES}} & \multicolumn{3}{c}{\textit{AVG}} \\
    \textbf{Model} & \texttt{AR} & \texttt{RU} & \texttt{ES} & \texttt{ZH} & \texttt{AR} & \texttt{RU} & \texttt{ES} & \texttt{ZH} & \texttt{KA} & \texttt{YO} & \texttt{MN} & \texttt{KA} & \texttt{YO} & \texttt{MN}  &  &  & \textbf{HRL} & \textbf{LRL} & \textbf{Overall} \\
    \midrule
    GPT \texttt{text-embedding-3-large} & 0.89&0.82&0.84&\textbf{0.92}&0.91&0.94&0.87&\textbf{0.87}&0.71&0.7&0.54&0.56&0.68&0.6&0.62&0.51 &\textbf{0.88}&0.62&0.75\\
    Cohere \texttt{Embed v3} & 0.84&0.87&0.83&0.88&0.9&\textbf{0.96}&\textbf{0.89}&0.83&0.75&0.73&0.54&0.58&0.74&0.64&0.65&0.59&0.88&\textbf{0.65}&0.76 \\
     \texttt{Mistral-embed} & \textbf{0.92}&0.88&0.82&0.85&0.92&0.86&0.86&0.83&0.72&0.7&0.56&0.53&0.68&0.61&0.63&0.53&0.87&0.62&0.74\\
    \texttt{SONAR} & 0.89&0.79&\textbf{0.85}&0.77&0.93&0.93&0.85&0.87&0.81&0.8&0.69&0.73&\textbf{0.79}&0.73&\textbf{0.69}&\textbf{0.62}&0.86&0.73&0.8\\
    \midrule
    \gpt & 0.8&0.72&0.65&0.8&0.86&0.91&0.86&0.79&0.61&0.57&0.26&0.47&0.43&0.31&0.38&0.4&0.8&0.43&0.61\\
    \gpto & 0.71&0.74&0.65&0.8&0.86&0.86&0.74&0.8&0.64&0.58&0.3&0.47&0.59&0.4&0.45&0.41&0.77&0.48&0.63\\
    \texttt{Command R} & 0.56&0.88&0.61&0.83&0.86&0.84&0.77&0.68&0.47&0.51&0.19&0.16&0.19&0.33&0.37&0.3&0.75&0.32&0.53\\
    \crplus & 0.59&0.56&0.65&0.7&0.91&0.91&0.76&0.74&0.34&0.39&0.04&0.41&0.43&0.26&0.15&0.4&0.73&0.3&0.51\\
    \texttt{Mistral 8x22b} & 0.25&0.59&0.53&0.67&0.84&0.94&0.74&0.77&0.51&0.4&0.08&0.46&0.52&0.5&0.33&0.46&0.67&0.41&0.54\\
    \texttt{Sonnet} & 0.7&0.75&0.61&0.8&0.84&0.89&0.7&0.69&0.64&0.62&0.41&0.56&0.58&0.55&0.53&0.47&0.75&0.55&0.65\\
    \texttt{Opus} & 0.6&\textbf{0.91}&0.69&0.83&0.88&0.9&0.83&0.76&0.66&0.54&0.2&0.49&0.7&0.53&0.33&0.49&0.8&0.49&0.65\\
    \llama & 0.6&\textbf{0.91}&0.69&0.83&0.88&0.9&0.83&0.76&0.66&0.54&0.2&0.49&0.7&0.53&0.33&0.49&0.8&0.49&0.65\\
    \midrule
    BLASER 2.0-QE & 0.9&0.89&0.85&0.78&\textbf{0.94}&0.92&0.87&0.87&\textbf{0.81}&\textbf{0.83}&\textbf{0.79}&\textbf{0.73}&0.78&\textbf{0.8}&0.68&0.58&0.88&\textbf{0.75}&\textbf{0.81}\\
    \bottomrule
    \bottomrule
    
  \end{tabular}
  }
  \vspace{-3mm}
  \caption{ROC-AUC results for severity hallucination ranking across HRL and LRL directions. \\ \textit{Bold values indicate the best performing prompt per model.}}
  \vspace{-4mm}
  \label{tab:ranking_results}
  \end{center}
\end{table*}

\section{Prompts}
\label{sec:app_prompts}

We used two types of CoTs: One based on the human guidelines for hallucination detection, and the other based on the severity level definition, that was readapted to each case. For \textbf{binary detection}, two CoTs were tested for three prompts.

\paragraph{Binary detection, Prompt1 - from \textit{G-Eval}} See {\autoref{fig:binary_prompt1}} 

\begin{figure}[ht]
    \centering
    \begin{tabular}{|p{0.9\linewidth}|}
    \hline
    \textit{System:}\\
    Human Evaluation of Machine Translation Systems: \\
    Hallucination Evaluation Criteria: Does the translated text contain information completely unrelated to the source text? \\
    - \textit{Hallucination}: there is hallucination. \\
    - \textit{No Hallucination}: there is no hallucination. \\ \\
        \textit{User:} \\
    Source Text: \texttt{src\_text}\\Translated Text: \texttt{mt\_text}\\    Does the translation contain hallucination? Answer (label ONLY: 'Hallucination' OR 'No Hallucination'):
    \\ \hline
    \end{tabular}
    \caption{\textbf{Binary detection Prompt 1} - from \textit{G-Eval}}
    \label{fig:binary_prompt1}
\end{figure}

\paragraph{Binary detection, Prompt2 - from \textit{G-Eval} with language precision} See {\autoref{fig:binary_prompt2}} 
\begin{figure}[ht]
    \centering
    \begin{tabular}{|p{0.9\linewidth}|}
    \hline
    \textit{System:}\\
    Instructions for Evaluating Machine Translation: \\
    You will be given a source text in \texttt{src\_lang} and a machine translated text in \texttt{tgt\_lang}. \\
    Your task is to identify if the machine translated text has hallucination or not. \\
    Please make sure you read and understand these instructions carefully. \\
    Please keep this document open while reviewing, and refer to it as needed. \\
    Evaluation Criteria: \\
    Hallucination: Does the translated text contain information completely unrelated to the source text? \\
    - \textit{Hallucination}: there is hallucination. \\
    - \textit{No Hallucination}: there is no hallucination. \\ \\
    
     \textit{User:} \\
    Source Text: \texttt{src\_text}\\Translated Text: \texttt{mt\_text}\\    Does the translation contain hallucination? Answer (label ONLY: 'Hallucination' OR 'No Hallucination'):
    \\ \hline
    \end{tabular}
    \caption{\textbf{Binary detection Prompt 2} - from \textit{G-Eval} with language precision}
    \label{fig:binary_prompt2}
\end{figure}

\paragraph{Binary detection, Prompt3 - Human designed prompt} See {\autoref{fig:binary_prompt3}} 

\begin{figure}[ht]
    \centering
    \begin{tabular}{|p{0.9\linewidth}|}
    \hline
    \textit{System:}\\
    Instructions for Evaluating Machine Translation: \\
    You will be given a source text in \texttt{src\_lang} and a machine translated text in \texttt{mt\_lang}. \\
    Your task is to identify if the machine translated text has hallucination or not.  Please make sure you read and understand these instructions carefully. Please keep this document open while reviewing, and refer to it as needed.\\
    \textbf{Definition of Hallucination:}
The translated text is considered a hallucination if it introduces information that is completely unrelated to the source text.

\textbf{Hallucination labels:}
\begin{itemize}
    \item \textbf{Hallucination:} there is hallucination.
    \item \textbf{No hallucination:} there is no hallucination.
\end{itemize}

     \textit{User:} \\
    Source Text: \texttt{src\_text}\\Translated Text: \texttt{mt\_text}\\  Provide exactly one of the following hallucination labels as your response. Do not include any additional text or explanation:
    \begin{itemize}
        \item \textit{Hallucination}
        \item \textit{No hallucination}:
    \end{itemize}
    \\ \hline
    \end{tabular}
    \caption{\textbf{Binary detection Prompt 3} - Human designed prompt}
    \label{fig:binary_prompt3}
\end{figure}

\paragraph{Binary detection, Chain of Thoughts} See {\autoref{fig:binary_cot1}} and {\autoref{fig:binary_cot2}} 

\begin{figure}[t]
    \centering
    \begin{tabular}{|p{0.9\linewidth}|}
    \hline
    \textbf{Evaluation Steps:} \\
    1. Read the source text and the translated text carefully. \\
    2. To decide whether the translated text contains hallucinations check if the source tokens "correspond" to erroneous target tokens. For each token answer: \\
    \begin{itemize}
        \item Does this source word fall into the common meaning category as this target word?
        \item Does this source word have a semantic connection with this target word?
        \item Can you try to come up with a reasonable theory on how this source word is associated with this target word?
    \end{itemize}
    3. If "no" to all the questions above, then hallucination 
    \\ \hline
    \end{tabular}
    \caption{\textbf{Binary detection - CoT1:} from \textit{HalOmi}'s human guidelines}
    \label{fig:binary_cot1}
\end{figure}

\begin{figure}[t]
    \centering
    \begin{tabular}{|p{0.9\linewidth}|}
    \hline
    \textbf{Evaluation Steps:} \\
    \begin{enumerate}
        \item Read the source text and the translated text carefully.
        \item Initialize a counter `n = 0` for the number of hallucinated words.
        \item To decide whether the translated text contains hallucinations check if the source tokens "correspond" to erroneous target tokens. For each token answer:
\begin{itemize}
        \item Does this source word fall into the common meaning category as this target word?
        \item Does this source word have a semantic connection with this target word?
        \item Can you try to come up with a reasonable theory on how this source word is associated with this target word?
        \item If "no" to all the questions above, then hallucination
    \end{itemize}
    \item After analyzing each word in the translated text:
    \begin{itemize}
        \item If `n == 0`, assign the label 'No hallucination'.
        \item If `n` is 1 or more, assign the label 'Hallucination'.'''
    \end{itemize}
    \end{enumerate}
    \\ \hline
    \end{tabular}
    \caption{\textbf{Binary detection - CoT2:} from \textit{HalOmi}'s human guidelines and counting strategy}
    \label{fig:binary_cot2}
\end{figure}

\section{LLMs experiments}
\label{sec:app_LLM_exp}

\subsection{LLMs hyperparameters}
For the evaluation of LLMs, we used \textit{LangChain} to ensure reproducibility of results, except for \llama that was ran locally. 
We set the \texttt{TEMPERATURE} to \texttt{0} for minimum randomness and the \texttt{MAX\_OUTPUT\_TOKEN} to \texttt{15} to avoid verbose.
All the experiments were zero-shot, with an exhaustive label (for example, \textit{['Hallucination', 'No Hallucination']} for \textbf{binary detection}). These choices showed the highest performances in previous research \cite{Kocmi2023} \cite{Wei2022}. 

\subsection{LLMs selection}
\label{sec:llm_choice}
We selected the following models for our evaluation: \texttt{GPT4-turbo}, widely adopted in both academic research and industrial applications due to its robust performance and versatility; \texttt{GPT4o}, the latest GPT model, optimised for better human-computer interaction; \texttt{Command-R}, known for its large context window, well-suited for tasks that require extended language understanding and generation; \texttt{Command R+}, an enhanced version of \texttt{Command R}, demonstrating strong performance in multilingual tasks, achieving impressive BLEU scores in benchmark datasets such as \href{https://txt.cohere.com/command-r-plus-microsoft-azure/}{FLoRES and WMT23}; \texttt{Mistral 8x22b}, currently the most performant open model from Mistral, excelling in various language tasks; \texttt{Claude Sonnet}, showing strong capabilities in multilingual tasks, similar to \texttt{Command R+}; \texttt{Claude Opus}, known as the "most intelligent" Claude model, offering advanced language understanding and generation capabilities; and \texttt{LLama3-70B}, the most capable openly available LLM from Meta, evaluated in its 70B size for comprehensive performance analysis. These models were chosen based on their demonstrated performance in various benchmarks and their potential to handle a wide range of language tasks effectively.



\clearpage
\section{Binary detection results}
\label{sec:binary_result}

\subsection{Validation results}
\label{sec:app_val_bin}
{\autoref{tab:val_results_binary}} provides MCC scores per LLM for each of the prompts and CoT variations evaluated on the validation set.
The most robust LLMs across prompt variations in the validation set, specifically Sonnet, \texttt{GPT4o}, and \llama, exhibit superior performance across language resource settings in the test set. This suggests that extensive prompt engineering might not be required for these models in the current task, as the performance using the optimal prompt from the validation set aligns with high performance on the test set.
\begin{table*}[!h]
 \footnotesize
 \begin{center}
 \resizebox{0.8\textwidth}{!}{%
  \begin{tabular}{lrr|rr|rrr||rr}
    \toprule
     & \multicolumn{2}{c}{\textit{Prompt1}} & \multicolumn{2}{c}{\textit{Prompt2}} & \multicolumn{3}{c}{\textit{Prompt3}} & \multicolumn{2}{c}{\textit{AVG}} \\
    \textbf{Model} & \textbf{no CoT} & \textbf{CoT1} & \textbf{no CoT} & \textbf{CoT1} & \textbf{no CoT} & \textbf{CoT1} & \textbf{CoT2} & \textbf{Mean} & \textbf{Std.} \\
    \bottomrule
    \midrule
    & \multicolumn{9}{c}{\textbf{Binary Detection (MCC)}} \\
    \midrule
    \gpt & 0.53 & 0.55 & \textbf{0.55} & 0.50 & 0.45 & 0.51 & 0.47 & 0.51 & 0.04 \\
    \gpto & 0.44 & 0.44 & \textbf{0.51} & 0.45 & 0.44 & 0.47 & 0.48 & 0.46 & 0.03 \\
    \texttt{Command R} & 0.43 & 0.37 & 0.54 & 0.47 & 0.51 & 0.53 & \textbf{0.55} & 0.49 & 0.07 \\
    \crplus & 0.72 & 0.72 & 0.57 & 0.69 & 0.54 & \textbf{0.72} & 0.64 & 0.66 & 0.08 \\
    \texttt{Mistral 8x22b} & 0.51 & 0.57 & 0.52 & 0.61 & \textbf{0.69} & 0.65 & 0.69 & 0.61 & 0.07 \\
    \texttt{Sonnet} & 0.67 & 0.68 & \textbf{0.69} & 0.68 & 0.68 & 0.69 & 0.68 & \textbf{0.68} & 0.01 \\
    \texttt{Opus} & 0.57 & 0.50 & 0.53 & 0.56 & \textbf{0.73} & 0.64 & 0.59 & 0.59 & 0.08 \\
    \llama & 0.74 & 0.76 & 0.74 & 0.72 & \textbf{0.81} & 0.79 & 0.79 & 0.76 & 0.03 \\
    \bottomrule
    \bottomrule
  \end{tabular}
  }
\caption{Validation results for binary hallucination detection across prompt variations. Bold values indicate the best performing prompt per model. In the case of ties, we favor shorter prompts without CoT.}
  \label{tab:val_results_binary}
 \end{center}
\end{table*}


\subsection{Test results}
\begin{figure*}[!h]
    \centering
    \includegraphics[width=\linewidth]{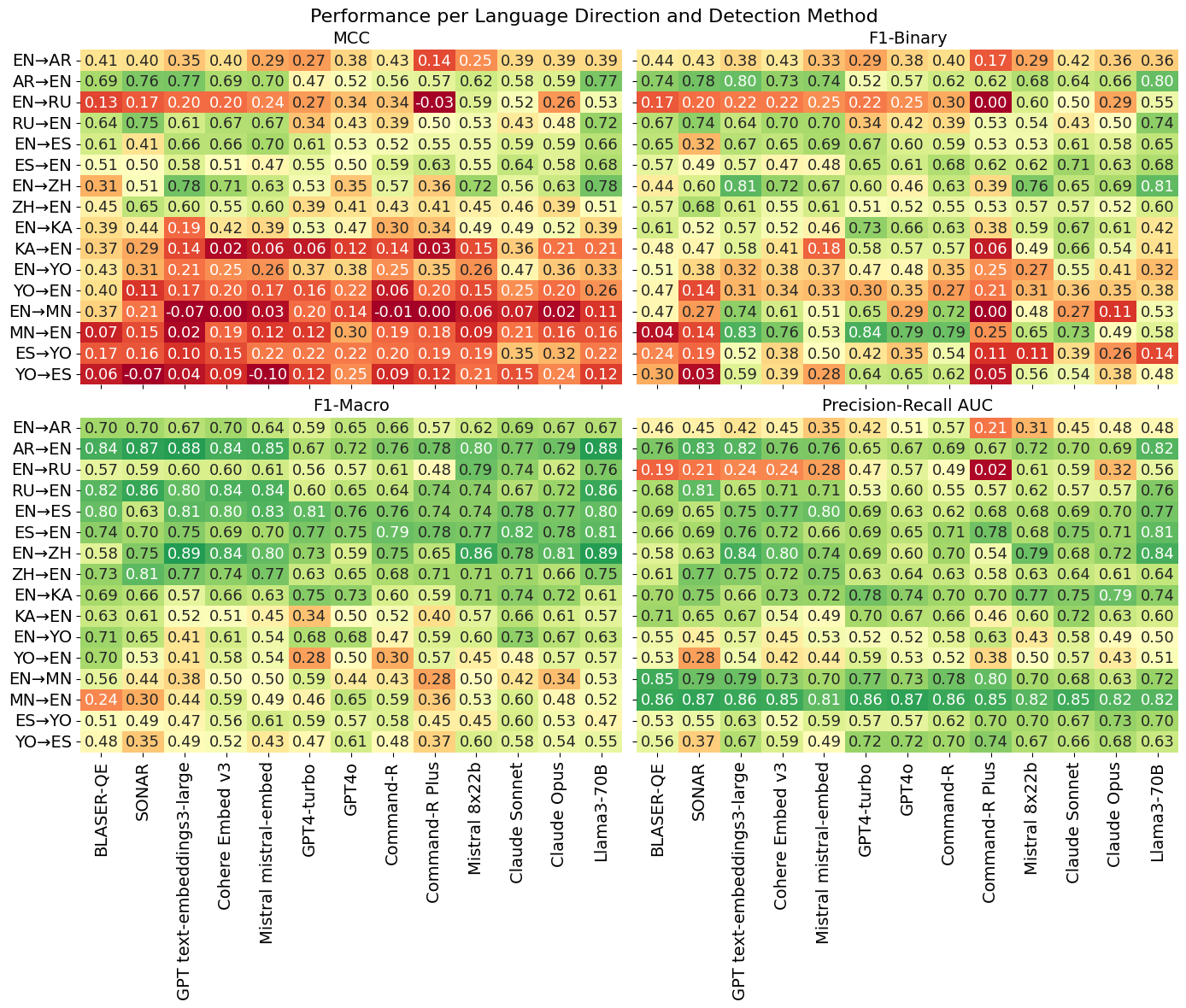}
    \caption{ MCC, binary F1, macro F1 and precision-recall AUC scores for hallucination binary detection across 16 translation directions per method. 
    }
    \label{fig:results_heatmaps}
\end{figure*}

\begin{figure*}[!h]
    \centering
    \includegraphics[width=\linewidth]{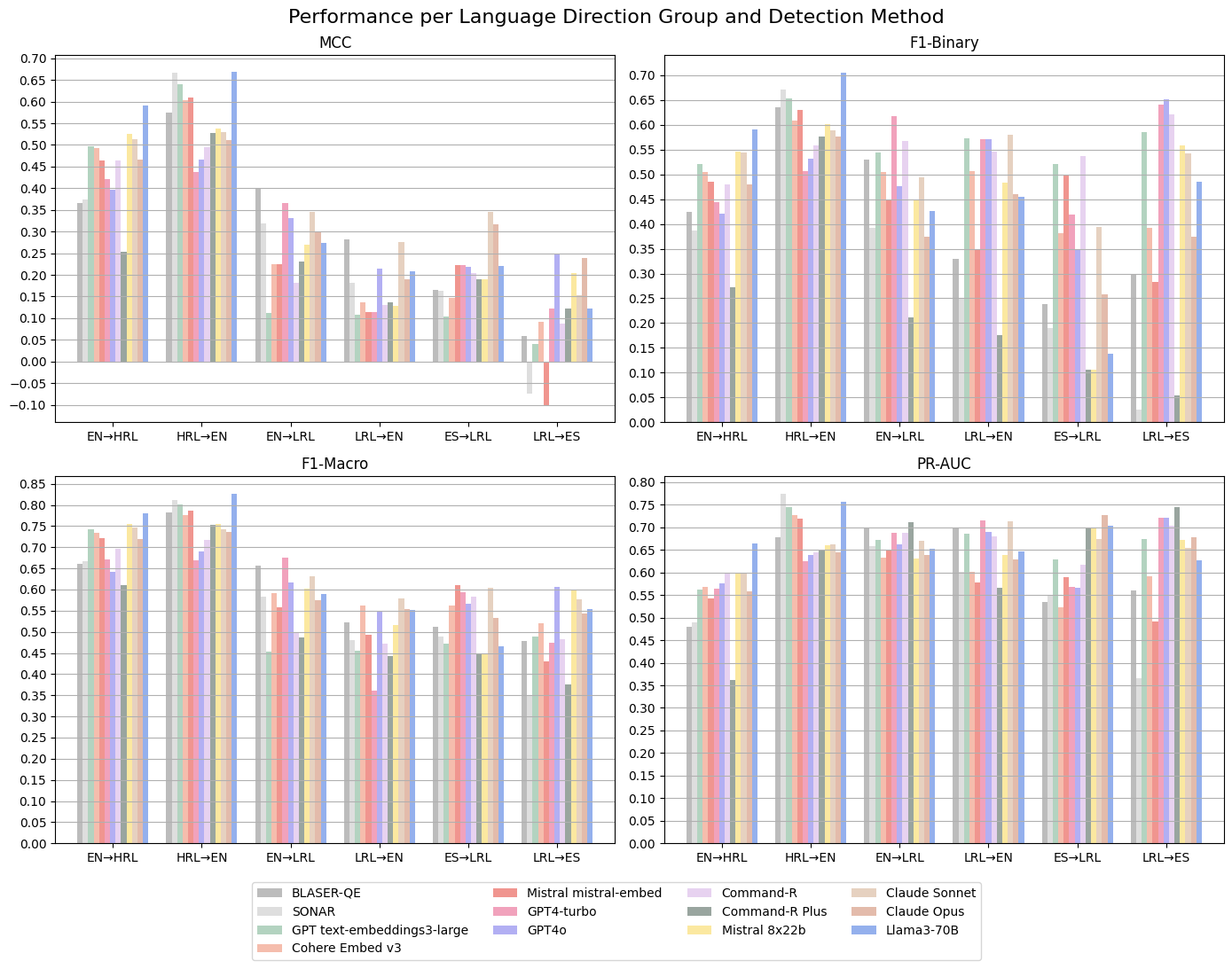}
    \caption{MCC, binary F1, macro F1 and precision-recall AUC average scores across high and low resource levels, for different directions. 
    }
    \label{fig:results_bar_agg}
\end{figure*}

\autoref{fig:results_heatmaps} and \autoref{fig:results_bar_agg} display the performances of the evaluated methods on the test set, grouped by translation directions. To account for the significant class imbalance, multiple metrics are employed to ensure a more comprehensive and unbiased analysis. These include the MCC, binary F1-score, macro F1-score, and precision-recall area under the curve (AUC).
The results indicate that the highest scores for HRLs are achieved in translations to English, whereas for LRLs, the highest scores are from translations originating in English or Spanish. Additionally, these findings underscore that no single model uniformly excels across all translation directions. Finally, the model rankings remain consistent across metrics in HRLs settings. However, there is greater variability in LRL scenarios, particularly for non-English centric translation directions. This variability is largely due to models like \texttt{BLASER-QE} and \texttt{SONAR}, exhibiting a high ratio of non-hallucination predictions, while others, such as \texttt{Command-R} and \texttt{GPT-embeddings}, show a stronger tendency towards hallucination predictions. 


\end{document}